\documentclass{article}
\usepackage{spconf,amsmath,amssymb,graphicx,hyperref}
\usepackage{booktabs, multirow}

\title{A Convolutional Neural Deferred Shader for Physics Based Rendering}
\name{Zhuo He\sthanks{Equal contribution.}, Yingdong Ru\footnotemark[1], Qianying Liu, Paul Henderson, Nicolas Pugeault}
\address{School of Computing Science, University of Glasgow\\
  Glasgow G12 8RZ, United Kingdom}
\begin{document}
%
\maketitle
%
\begin{abstract}
Recent advances in neural rendering have achieved impressive results on photorealistic shading and relighting, by using a multilayer perceptron (MLP) as a regression model to learn the rendering equation from a real-world dataset. Such methods show promise for photorealistically relighting real-world objects, which is difficult to classical rendering, as there is no easy-obtained material ground truth. 
However, significant challenges still remain---the dense connections in MLPs result in a large number of parameters, which requires high computation resources, complicating the training, and reducing performance during rendering. Data driven approaches require large amounts of training data for generalization; unbalanced data might bias the model to ignore the unusual illumination conditions, e.g.~dark scenes.
This paper introduces \textit{pbnds+}: a novel physics-based neural deferred shading pipeline utilizing convolution neural networks to decrease the parameters and improve the performance in shading and relighting tasks; Energy regularization is also proposed to restrict the model reflection during dark illumination. Extensive experiments demonstrate that our approach outperforms classical baselines, a state-of-the-art neural shading model, and a diffusion-based method.
\end{abstract}
\renewcommand{\paragraph}[1]{\par\noindent\textbf{#1}~}

\section{Introduction}
Photo-realism is a fundamental goal of computer graphics~\cite{newell_progression_1977}. It relies on accurately modeling light–surface interactions through complex light-transport processes. Recent progress has been made in illumination representation, material appearance modeling, and high-fidelity geometric reconstruction \cite{deering_triangle_1988,dick2009efficient,guarini2024pbr,walter_microfacet_2007}; this allows neural network-based \textit{inverse approaches} to estimate materials and illumination of real-world scene, then to use \textit{forward approaches} to render real-world objects photorealistically with various illumination conditions. However, approximations are required for complex calculations of real-world rendering, and inverse approaches are usually tied to specific rendering models, which restricts the rendering performance and applications.

Recent advances tackle this problem through a fully data-driven approach, where a multilayer perceptron (MLP) is used to regress a rendering function from estimated materials and illumination to an image of a real-world object~\cite{he2025beyond}. This greatly improves the rendering performance compared to classical rendering models. However, the MLP is a dense, high-parameter architecture, influencing the speed and performance of rendering. Inspired by convolution neural networks~(CNN)~\cite{krizhevsky2012imagenet}, we propose a convolutional neural deferred shader named \textbf{pbnds+} to overcome drawbacks of the existing work; we also introduce an energy regularization operation in our training process, to improve the rendering results in dark illumination. Our contributions are as follows:
\begin{enumerate}
  \item We propose a novel convolution-based neural deferred shading pipeline that renders scenes with PBR textures (albedo, roughness, specular) and illumination (HDRI light map) to photo-realistic images.
  \item We develop an energy regularization to rectify our data-driven rendering, especially for the rendering in dark illumination conditions.
  \item Extensive experiments are conducted to compare our approach with existing shading models for real-world objects rendering on multiple datasets, demonstrating its superior performance.
\end{enumerate}

\begin{figure*}[t]
    \centering
    \includegraphics[width=1.0\textwidth]{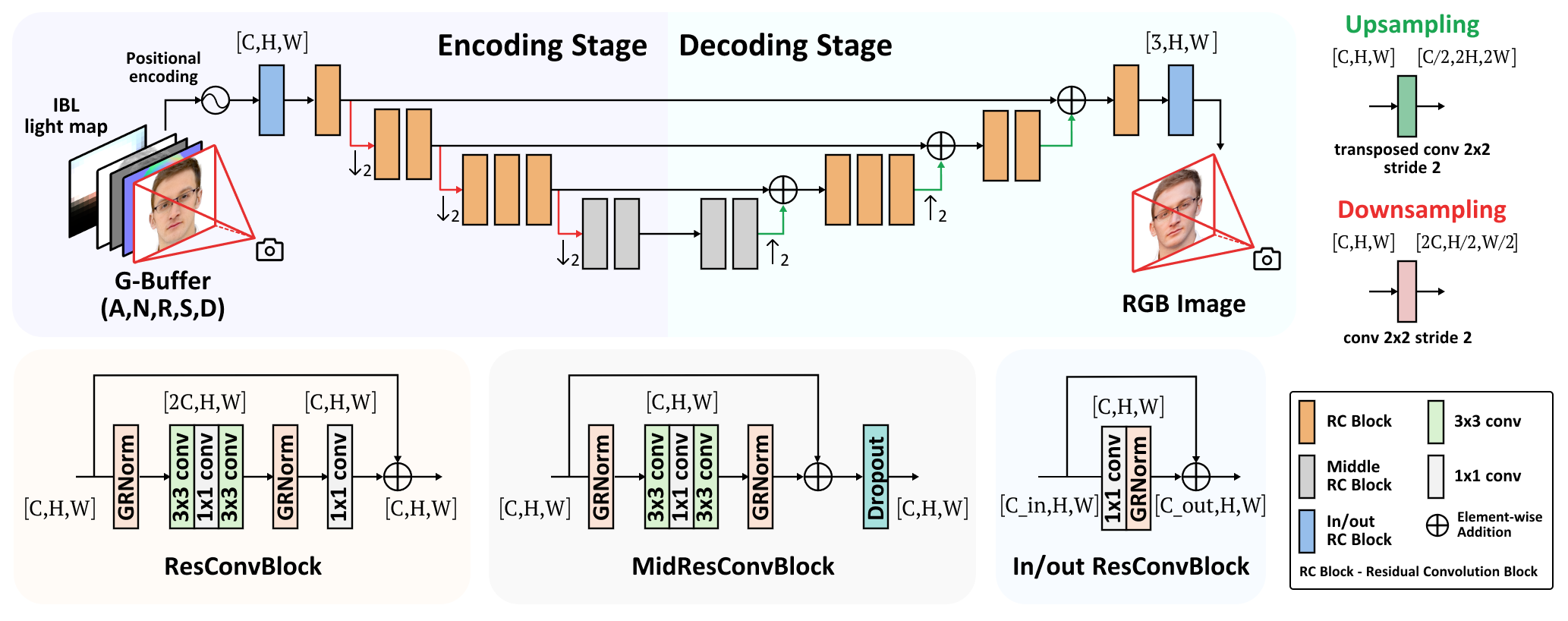}
    \caption{{The overall pipeline of our convolutional neural deferred shading. Given an input image we estimate PBR textures (A: albedo, N: Normal, S: specular, R: roughness, D: depth), IBL lightmap, and field of view, using pre-trained models. Then the estimated data are used to train the convolutional neural deferred shader}}
    \label{fig:fig01}
\end{figure*}

\section{Related Works}
\paragraph{Neural Rendering}
Unlike traditional rendering pipelines that require explicit geometry and material modeling, neural rendering learns implicit or hybrid scene representations directly from data, enabling high-quality synthesis under challenging conditions. For instance, Neural Radiance Fields (NeRF) \cite{mildenhall_nerf_2020} represent a single 3D scene as an implicit radiance field function optimized from multi-view images, allowing to render novel views in photorealistic quality. Building upon this foundation, following works~\cite{barron2021mip}~\cite{muller2022instant}have focused on improving efficiency, scalability, and realism. 
More recently, 3D Gaussian splatting (3DGS)~\cite{kerbl_3d_2023,szymanowicz_splatter_2024} introduced an explicit point-based representation that allows real-time rendering, and subsequent extensions explored differentiable shading and inverse rendering~\cite{liang2024gs, chen2024gi}. Both of these approaches rely on overfitting to specific scenes and struggle to generalize, as they entangle geometry, material, and illumination into a single representation. In contrast, our method builds on the standard deferred rendering pipeline, explicitly learning the shading process, which enables rendering of arbitrary scenes while producing photorealistic results.  
\vspace{0.5em}

\paragraph{Photorealistic Relighting}
Relighting aims to re-render a scene or subject under novel illumination conditions while preserving its geometry and material appearance. Traditional methods relied on explicit geometry and reflectance capture, often requiring controlled environments or specialized devices. While accurate, these approaches were difficult to scale to casual or in-the-wild scenarios. However, neural rendering and differentiable inverse rendering allow relighting from sparse or unstructured inputs, often without requiring full geometry or reflectance measurements~\cite{sun2019single,he2025beyond}~\cite{munkberg2022extracting}~\cite{liang2024gs}~\cite{jin2024neural}. 
However, existing methods either overfit to individual scenes and struggle to generalize, or introduce randomness into the rendering process, making it difficult to maintain the identity. Our approach extends existing MLP-based neural deferred shading pipeline with convolution neural network, improving the training speed and quality of shading results.

\section{Methods}
\paragraph{Problem setting} Image rendering models the interaction between light and surfaces~\cite{kim_switchlight_2024}, aiming to solve the \textit{rendering equation}:
\begin{equation}
    L_o(\mathbf{v}) = \int_{\Omega} F(\mathbf{v},\mathbf{l})L_i(\mathbf{l}) \langle \mathbf{n} \cdot \mathbf{l} \rangle d\mathbf{l}
    \label{eqn:eqn01}
\end{equation}
where $L_o(\mathbf{v})$ is the outbound radiance leaving in direction $\mathbf{v}$; it is the integral of the incident light $L_i(\mathbf{l})$ from every possible direction $\mathbf{l}$ across the hemisphere $\Omega$, centered around the surface normal $\mathbf{n}$. $F(\mathbf{v}, \mathbf{l})$ is the Bidirectional Reflectance Distribution Function (BRDF) describing how the surface reflects light. 
In this task, our goal is to use a convolutional neural network (CNN) to regress output of the rendering equation \ref{eqn:eqn01}. Specifically, we define the following process:
\begin{equation}
    L_o(\mathbf{v}) = \int_{\Omega} f_{\theta}(\mathbf{a}, \mathbf{n}, \mathbf{s}, \mathbf{r}, \mathbf{v}, L_{i}(\mathbf{l}) \langle \mathbf{n} \cdot \mathbf{l} \rangle) d\mathbf{l}, \mathbf{l} \in \Omega
    \label{eqn:eqn02}
\end{equation}
where $L_i(\mathbf{l}) \langle \mathbf{n} \cdot \mathbf{l} \rangle$ represents the inbound light from upper hemisphere $\Omega^+$ similar to equation~\ref{eqn:eqn01}. Our neural shader first predicts the outbound light contribution from each sampled inbound direction, then averages those predictions to approximate the hemispherical integral. Thus the neural shader learns the shading integral directly from the reconstruction loss:
\begin{equation}
    \mathcal{L}_{\mathrm{rec}}\;=\;\frac{1}{B}\sum_{i=1}^B\bigl\|\,\hat I_i(\mathbf{x}) \;-\; I^{\mathrm{gt}}_i(\mathbf{x})\bigr\|_1.
    \label{eqn:eqn03}
\end{equation}

\begin{figure}[h]
    \centering
    \includegraphics[width=0.48\textwidth]{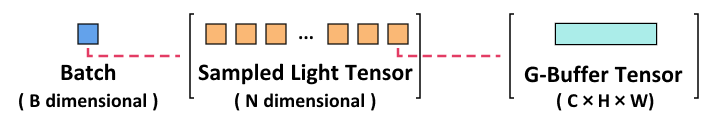}
    \caption{Structure of input tensor $X_{G}$. As 2D convolution only access three dimensional tensors, we combine the batch and light dimension for compatibility, thus each batch instance indicates the contribution of single sampled light ray.}
    \label{fig:fig02}
\end{figure}

\paragraph{Convolutional shading process} To make the neural rendering process be compatible with convolution neural network, we define the following operations to implement shading through convolution calculation. As the illustration of figure~\ref{fig:fig02}, the input tensor $X_{G}$ is defined by $\mathbb{R}^{B \times N \times C \times H \times W}$, where $B,N,C,H,W$ indicate the size of batch size, sampled incident light rays, input channels, height and width. During shading, $X_{in}$ is reshaped firstly to combine the dimension of batch and sampled light, the G-Buffer tensor formed by other dimensions $[C \times H \times W]$ are treated as a feature map corresponding to each sampled light for single batch instance. For each sampled incident light, CNN will regress the shading result of the whole feature map lit by an individual incident light, the overall shading result for one batch instance is a weighted sum among all sampled incident lights.

As the illustration of figure~\ref{fig:fig01}, our convolutional neural deferred shader implements above process through a UNet-like structure~\cite{ronneberger2015u}, it follows the encoder-decoder framework which comprehends information from input features during encoding stage via several down-sampling and feature extraction operations, then recover the target feature during decoding stage leveraging several up-sampling and feature regression operations, skip connections are enforced to help establish the identity from encoder to decoder. We employ the ResConvBlock to extract or regress features in down-sampling and up-sampling process, and the MidResCovBlock in bottleneck which adds the dropout layer to alleviate overfitting, improving model robustness. We utilize convolution layer with $2 \times 2$ kernel size and 2 stride for down-sampling, and transposed convolution layer with same configuration for up-sampling. All input values are encoded through positional encoding for utilizing the information of frequency domain and normalization.
\vspace{0.3em}

\paragraph{Energy regularization} The data-driven shading approach requires diverse data to generalize, However, when incident illumination is extremely low---an edge case rarely sampled during training, the model often fails to predict the correspondingly dark appearance, since it has never learned to map near‐zero incoming light to near‐zero output. To address this, we introduce an energy regularization loss (See equation~\ref{eqn:eqn04}): during training, we stochastically replace the environment map with an all‐zero “dark” HDRI input according to a Bernoulli gate, and explicitly penalize any nonzero shading output. A single hyperparameter $P_{dark}$ is used to control the probability of using this all‐dark illumination, ensuring that the network learns to produce truly dark renderings whenever incident light vanishes.

\begin{equation}
\begin{gathered}
    z_i \sim \mathrm{Bernoulli}(p_{0}), \quad i=1,\dots,B \\[6pt]
    \tilde L_{\mathrm{env},i}(\omega) = (1 - z_i)\,L_{\mathrm{env},i}(\omega), \\[6pt]
    \mathcal{L}_{\mathrm{zero}} = \frac{1}{B}\sum_{i=1}^B z_i \,\bigl\lVert \hat I_i(\mathbf{x})\bigr\rVert_2^2.
\end{gathered}
\label{eqn:eqn04}
\end{equation}


The training objective is a weighted sum of the reconstruction loss $\mathcal{L}_{\mathrm{rec}}$ and our energy regularization term $\mathcal{L}_{\mathrm{zero}}$:

\begin{equation}
\mathcal{L} = \mathcal{L}_{\mathrm{rec}} + \lambda_{\mathrm{zero}}\,\mathcal{L}_{\mathrm{zero}}.
\label{eqn:total_loss}
\end{equation}

\begin{table*}[tb]
  \centering
  \caption{Quantitative evaluation for shading/relighting experiments}
  \resizebox{\textwidth}{!}{
  \begin{tabular}{@{}clcccccccccc@{}}
    \toprule
      & & \multicolumn{2}{c}{\small MSE $\downarrow$} & \multicolumn{2}{c}{\small PSNR $\uparrow$} & \multicolumn{2}{c}{\small SSIM $\uparrow$} & \multicolumn{2}{c}{\small LPIPS $\downarrow$} & \multicolumn{2}{c}{\small FID $\downarrow$} \\
      \cmidrule(r){3-4} \cmidrule(r){5-6} \cmidrule(r){7-8} \cmidrule(r){9-10} \cmidrule{11-12} 
      & & \small FP & \small CP & \small FP & \small CP & \small FP & \small CP & \small FP & \small CP & \small FP & \small CP \\
    \midrule
      \multirow{4}{*}{\centering \textbf{SD}}
        & Blinn-Phong & 0.0938 & 0.0944 & 10.8205 & 11.9432 & 0.4887 & 0.5543 & 0.2311 & 0.2732 & 0.5436 & 0.4693 \\
        & GGX         & 0.1111 & 0.0855 & 10.0108 & 12.3583 & 0.4171 & 0.3021 & 0.2895 & 0.2130 & 0.7608 & 0.5422 \\
        & NDS (Overfitted) & 0.0036 & 0.0043 & \textbf{29.7414} & \textbf{30.5312} & 0.8850 & \textbf{0.9640} & 0.0614 & 0.0573 & 0.1792 & 0.1003 \\
        & Neural Gaffer & \textbf{0.0028} & 0.0032 & 27.6352 & 27.7353 & 0.7422 & 0.7628 & 0.0355 & 0.0367 & 0.0828 & \textbf{0.0720} \\
        & PBNDS & 0.0047 & 0.0036 & 24.1229 & 29.4328 & 0.8948 & 0.9264 & 0.0444 & 0.0342 & 0.0976 & 0.1128 \\
        & PBNDS+ (Ours) & 0.0044 & \textbf{0.0026} & 29.0071 & 28.3211 & \textbf{0.9312} & 0.9364 & \textbf{0.0305} & \textbf{0.0258} & \textbf{0.0574} & 0.0783 \\
    \hline
    \rule{0pt}{8pt}
      \multirow{3}{*}{\textbf{RE}}
        & Blinn-Phong   & - & - & - & - & - & - & - & - & 0.3312 & 0.1633 \\
        & GGX           & - & - & - & - & - & - & - & - & 0.5561 & 0.1624 \\
        & Neural Gaffer & - & - & - & - & - & - & - & - & 0.1044 & 0.1152 \\
        & PBNDS         & - & - & - & - & - & - & - & - & 0.0903 & \textbf{0.0820} \\
        & PBNDS+ (Ours) & - & - & - & - & - & - & - & - & \textbf{0.0868} & 0.0948 \\
    \bottomrule
  \end{tabular}}
  \label{tab:tab01}
  \parbox{0.9\linewidth}{\small \textbf{SD:} Shading experiment; \textbf{RE:} Relighting experiment; \textbf{FP:} FFHQPBR dataset; \textbf{CP:} CelebAPBR dataset}
\end{table*}

\section{Experiments}
We conducted comprehensive experiments to evaluate the performance of our convolutional neural deferred shading approach, comparing the quality of shading and relighting results against five baseline models, including two classical shading models (the empirical Blinn-Phong model and physics-based GGX model); two recent learning-based model, neural deferred shader~\cite{worchel_multi-view_2022}, physics-based neural deferred shader~\cite{he2025beyond}; a state-of-art diffusion-based model, neural grapher~\cite{jin2024neural}. All experiments in this work are trained on FFHQ-PBR and CelebaA-PBR dataset, two high-quality human facial datasets with estimated material feature maps and illumination. To quantitatively evaluate the performance of shading experiment, we use Learned Perceptual Image Patch Similarity (LPIPS) \cite{zhang_unreasonable_2018} and Fréchet inception distance (FID) \cite{heusel_gans_2017} as metrics; lower LPIPS and FID scores indicate better realism comparing ground truth and real world data. We also measure the Peak signal-to-noise ratio (PSNR) and Structural similarity index measure (SSIM), where higher PSNR or SSIM score indicates better reconstruction performance. As there is no paired ground-truth data for the relighting experiment, we only use FID score for evaluation.

\begin{figure}[h]
    \centering
    \includegraphics[width=0.48\textwidth]{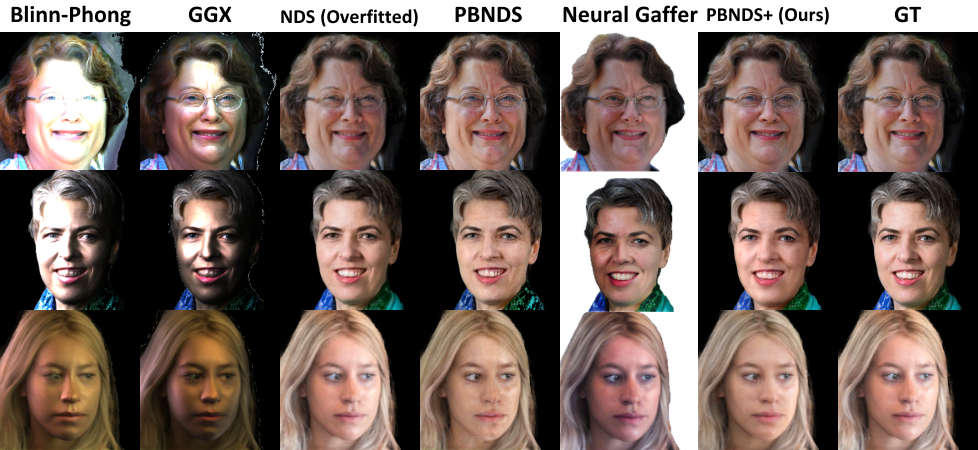}
    \caption{Quality comparison between different shading models.}
    \label{fig:fig03}
\end{figure}

\paragraph{Shading experiment}
For shading experiment, we visually compare shading outputs from various models. figure~\ref{fig:fig03} shows that our convolutional neural deferred shader (PBNDS+) delivers the most photorealistic reconstructions, faithfully reproducing hue and reflections from estimated materials and lighting. In contrast, Blinn–Phong’s empirical formulation cannot achieve true photorealism, and GGX’s microfacet theory requires manual energy-conservation tuning to match real illumination. We also benchmark against a reconstruction style neural deferred shading (NDS)~\cite{worchel_multi-view_2022}, a generalizable style physics-based neural deferred shading (PBNDS)~\cite{he2025beyond} and a diffusion based relighting model~\cite{jin2024neural} (See table~\ref{tab:tab01}): despite NDS overfitting per scene by ingesting scene-specific point coordinates, PBNDS introduces sampled training method to deal with dense parameterization, Neural Gaffer model the shading in generative way losing the physic prior such as reflection angle for accurate guidance. PBNDS+—being efficiently model scene-agnostic—matches its quality while generalizing across inputs.

\begin{figure}[h]
    \centering
    \includegraphics[width=0.48\textwidth]{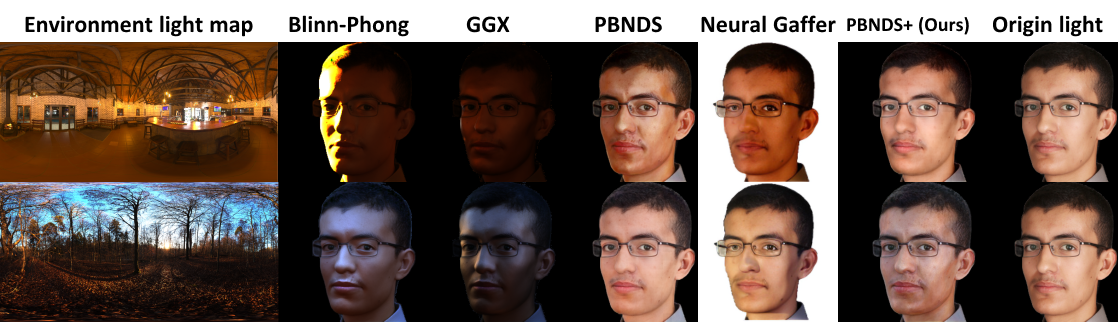}
    \caption{Comparing the result of energy regularization.}
    \label{fig:fig04}
\end{figure}

\paragraph{Relighting experiment} We evaluate our model’s relighting performance by swapping in real-world HDRI maps at render time. Using the same network trained on two facial datasets, we replace the estimated illumination with captured HDR environments and compare the results against Blinn–Phong, GGX, the original PBNDS, and Neural Gaffer renderings (see figure \ref{fig:fig04}). Our enhanced PBNDS+ shader accurately reproduces complex light–surface interactions under previously unseen lighting: it captures fine details in shadowed regions that classical models miss and avoids the unnatural hard shadow edges produced by Neural Gaffer’s generative relighting. Unlike conventional methods that rely on estimated illumination, our approach generalizes reliably to real HDRI inputs—even those outside the training distribution—delivering high-fidelity relighting across diverse environments.

We also compare the relighting performance of relighting quantitatively (See table~\ref{tab:tab01}), our PBNDS+ excel other models on FFHQPBR dataset and achieving competitive result to the previous PBNDS model on CelebaAPBR dataset.

\begin{figure}[h]
    \centering
    \includegraphics[width=0.44\textwidth]{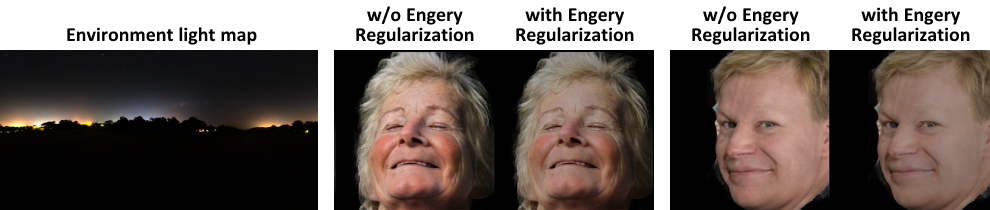}
    \caption{Comparing the result of relighting experiment.}
    \label{fig:fig05}
\end{figure}

\paragraph{Ablation study} We performed an ablation study on our covolutional neural deferred shading pipeline, evaluating effectiveness of our energy regularization loss. Figure~\ref{fig:fig05} illustrates the rectified result under a very dark illumination condition, currently the result can correctly reflect the scene illumination condition.

\section{Conclusion}
We introduce a novel extension to the physics-based neural deferred shading pipeline that delivers superior performance over existing approaches. Our enhanced model enables high-quality rendering and relighting—outperforming baseline methods—while effectively resolving artifacts in low‐illumination regions. Although our current formulation does not explicitly account for volumetric scattering or refraction, and thus cannot yet handle transparent materials, we plan to incorporate these effects in future work to achieve truly comprehensive shading.

\bibliographystyle{IEEEbib}
\bibliography{pbnds+}

\end{document}